\documentclass{article}

\usepackage[english]{babel}

\usepackage{authblk}
\date{} 

\providecommand{\keywords}[1]{\textbf{Keywords:} #1}

\usepackage[top=1in, bottom=1in, left=1in, right=1in]{geometry} 

\usepackage{lineno}
\usepackage{amsmath}
\usepackage{graphicx}
\usepackage[colorlinks=true, allcolors=blue]{hyperref}
\usepackage{abstract}
\usepackage{setspace}
\usepackage[square,numbers,sort&compress]{natbib}

\usepackage[table]{xcolor}
\usepackage{booktabs}

\usepackage{doi}

\title{Learning Latent Hardening (LLH): Enhancing Deep Learning with Domain Knowledge for Material Inverse Problems}

\author[1]{Qinyi Tian}
\author[2]{Winston Lindqwister}
\author[1]{Manolis Veveakis} 
\author[1,*]{Laura E. Dalton} 

\affil[1]{Duke University, Department of Civil \& Environmental Engineering, Durham, NC, USA}
\affil[2]{Delft University of Technology, Department of Civil Engineering \& Geosciences, Delft, South Holland, Netherlands}

\affil[*]{Corresponding author email: laura.dalton@duke.edu}

\begin{document}

\maketitle
\doublespacing
\begin{abstract}

Advancements in deep learning and machine learning have improved the ability to model complex, nonlinear relationships, such as those encountered in complex material inverse problems. However, the effectiveness of these methods often depends on large datasets, which are not always available. In this study, the incorporation of domain-specific knowledge of the mechanical behavior of material microstructures is investigated to evaluate the impact on the predictive performance of the models in data-scarce scenarios. To overcome data limitations, a two-step framework, Learning Latent Hardening (LLH), is proposed. In the first step of LLH, a Deep Neural Network is employed to reconstruct full stress-strain curves from randomly selected portions of the stress-strain curves to capture the latent mechanical response of a material based on key microstructural features. In the second step of LLH, the results of the reconstructed stress-strain curves are leveraged to predict key microstructural features of porous materials. The performance of six deep learning and/or machine learning models trained with and without domain knowledge are compared: Convolutional Neural Networks, Deep Neural Networks, Extreme Gradient Boosting, K-Nearest Neighbors, Long Short-Term Memory, and Random Forest. The results from the models with domain-specific information consistently achieved higher $R^2$ values compared to models without prior knowledge. Models without domain knowledge missed critical patterns linking stress-strain behavior to microstructural changes, whereas domain-informed models better identified essential stress-strain features predictive of microstructure. These findings highlight the importance of integrating domain-specific knowledge with deep learning to achieve accurate outcomes in materials science.

\end{abstract}

\vspace{5mm}

\newpage
\doublespacing
\section{Introduction}
\begin{flushleft}

Porous materials play a vital role in daily life. From the bones in human bodies to the geological materials below the Earth's surface, the distinct mechanical properties of each material are inherently linked to the internal microstructures. However, establishing a clear understanding of the relationship between mechanical behavior and heterogeneous microstructures remains a significant challenge. The analysis of porous materials is further complicated by subtle variations in microstructures, making predicting properties, such as mechanical strength, particularly difficult \cite{LESUEUR2021102061}. In engineering design, forward problems can be used to predict structural responses from known parameters while inverse problems infer material properties from observed behavior. This approach is particularly useful for analyzing porous materials with complex microstructures, aligning with established inverse problem methodologies \cite{gallet2022}.

In the 1990s, researchers investigated material inversion through mathematical modeling by utilizing a modified Levenberg-Marquardt method \cite{https://doi.org/10.1002/nme.1620331004}. This technique aimed to capture the latent discrepancies between the finite element method (FEM) and unknown material parameters. By minimizing these differences, the method enabled the determination of the elastic properties of the material interfaces \cite{https://doi.org/10.1002/nme.1620331004}. Traditional methods, such as destructive mechanical strength testing and finite element analysis (FEA), have been effective tools to characterize mechanical behavior. However, both techniques can be inefficient and difficult to generalize, particularly when applied to large-scale or complex systems where variations in microstructure influence the mechanical response in nontrivial ways \cite{rao2010finite}. 

To overcome these challenges, researchers from many scientific disciplines are turning to deep learning (DL) and machine learning (ML) methods. Interpretable ML models have been shown to improve predicting material properties. For example, a convolutional neural network (CNN) was developed to link microstructural evolution and mechanical behavior in dual-phase steel with an accuracy of  94 \% while improving computational efficiency. It is important to note that CNNs are a specialized type of artificial neural network (ANN) designed to process spatial data, such as images or microstructures, using convolutional layers, whereas ANNs more broadly refer to networks that learn patterns in data without explicit spatial processing. Additionally, ML encompasses a wide range of algorithms that learn from data, while DL specifically refers to ML methods that utilize multi-layered neural networks to model complex patterns and representations. By either emulating model behavior for Bayesian calibration or directly predicting material parameters, these ANN-based approaches offer efficient solutions for inverse modeling in engineering applications \cite{POGORELKO2024108912}. DL models have been applied to subsurface resistivity estimation which has enabled real-time predictions and bypassed computational limitations of traditional gradient-based methods \cite{puzyrev2019deep}. Recent studies also highlight the role of uncertainty quantification to study microstructures by addressing both forward and inverse problems in process–structure and structure–property relationships \cite{Billah2024}. Hybrid DL architectures with integrated dimensionality reduction and adjoint-based optimization have been used to efficiently predict spatial-temporal $\text{CO}_2$ saturation fields which has improved subsurface monitoring while significantly reducing computational costs \cite{https://doi.org/10.1029/2021WR031041}. Similarly, ML-driven methods have improved four dimensional (4D) printing by enabling accurate prediction and design of complex architectures, using Residual Neural Network (ResNet) for forward modeling and evolutionary algorithms for inverse design \cite{JIN2024102373}.

The case of inverse problems is a new frontier to leverage the capabilities of DL to efficiently process large datasets. ML based inverse methods have been developed to extract mechanical properties of heterogeneous membranes from full-field strain distributions, achieving speed improvements compared to traditional inverse FEA approaches \cite{ZHANG2022104134}. Similarly, Physics-informed Information Field Theory (PIFT) has been introduced to integrate physical laws with measurement data while remaining independent of numerical discretization, enabling robust uncertainty quantification \cite{ALBERTS2023112100}. More recently, the FlowPaths numerical inverse method, which employs a graph-theoretical approach, was developed to estimate hydraulic conductivity fields in porous materials using specific discharge data \cite{mont2024inverse}. By enhancing numerical stability and robustness, DL/ML approaches may contribute to more reliable structural inferences in engineering applications. 

The objective of the present study is to solve the inverse problem from recent work \cite{Lindqwister2023} where DL was used to predict the stress-strain curves using a large dataset of high resolution, X-ray micro-computed tomography (CT) scans of porous materials \cite{peloquin2023neural}. 
While most models focus on predicting material properties from structure, the objective of the present work is to reverse the process by predicting the underlying microstructure based on mechanical behavior. This inverse approach contributes to design, as it enables the reconstruction of material architectures from observed mechanical responses. To test this objective, Learning Latent Hardening (LLH), a customized DNN, was introduced to address the inverse problem by reconstructing the full stress-strain curve before predicting the microstructure. LLH leverages DL to infer missing mechanical responses, ensuring the material characterization is based on a complete representation of stress-strain behavior rather than fragmented or partial data. By first reconstructing the stress-strain curve, LLH provides a more comprehensive input for microstructural prediction, capturing latent hardening effects otherwise lost in incomplete datasets. Through this framework, the potential of DL/ML in solving inverse problems and the impact of incorporating domain knowledge on predictive performance are explored.

\end{flushleft}

\section{Methods}
\begin{flushleft}

This study introduceS a systematic framework called Learning Latent Hardening (LLH) to predict microstructural characteristics and reconstruct full stress-strain curves from partial data using DL/ML techniques. The LLH workflow consists of two main stages: (1) a Deep Neural Network (DNN)employed to reconstruct the full stress-strain curve based on incomplete stress-strain data to capture the latent mechanical response of the material and (2) the reconstructed stress-strain curves were then utilized to predict the underlying microstructural features to enable an effective inverse mapping between mechanical properties and material microstructure. A carefully prepared dataset was used alongside five DL/ML methods - Convolutional Neural Networks (CNN), K-Nearest Neighbors (KNN), Long Short-Term Memory (LSTM), Random Forest (RF), and Extreme Gradient Boosting (XGBoost) - to test the feasibility of applying DL/ML to solve the inverse problem. In this study, incorporating domain-specific knowledge to enhance model accuracy and improve the reliability of key microstructural feature predictions was also examined. Model performance was evaluated using established metrics to ensure a thorough assessment of predictive capabilities.
\end{flushleft}

\subsection{Data Preprocessing}
\begin{flushleft}

The dataset for this study was from the forward model developed by Lindqwister et al. \cite{Lindqwister2023}, which provided outputs for 35 features derived from multiple X-ray micro-computed tomography (CT) scans of various geological, cement-based, and wood materials. Each of the 35 features corresponds to a microstructural characteristic of the samples processed using an open-source FEM tool, Multiphysics Object Oriented Simulation Environment (MOOSE) \cite{Lindsay2022, Poulet2016}, and open source MorphoLibJ plugin in ImageJ software \cite{Legland2016, schneider2012nih}. 
In addition to these 35 microstructural characteristics, the corresponding stress-strain data were used as pre-information to train each DL/ML models. Before utilizing the stress-strain data for model training, a preprocessing step was applied where a portion of each stress-strain curve was randomly masked (as shown in \textbf{Figure 1}). This masking process removed parts of the stress-strain curve obtained from the forward model and created incomplete inputs that required the first step of the LLH model to infer the missing mechanical response. The masked regions were randomly chosen along the strain axis (i.e., x-axis) to ensure variability across the dataset and prevent the model from learning trivial patterns. The percentage of missing data was varied between the stress-strain curve data to introduce robustness in the learning process and to allow the model to generalize effectively between different material behaviors. Once the masking was applied, the modified dataset was structured into input-output pairs, where the masked stress-strain curves served as inputs and the corresponding unmasked data provided the reference output. This structured preparation was used to ensure the dataset retained essential mechanical information while enhancing the ability of the model to extract latent relationships from incomplete stress-strain data.

\begin{figure}[!htb]
\centering
\includegraphics[width=1\textwidth]{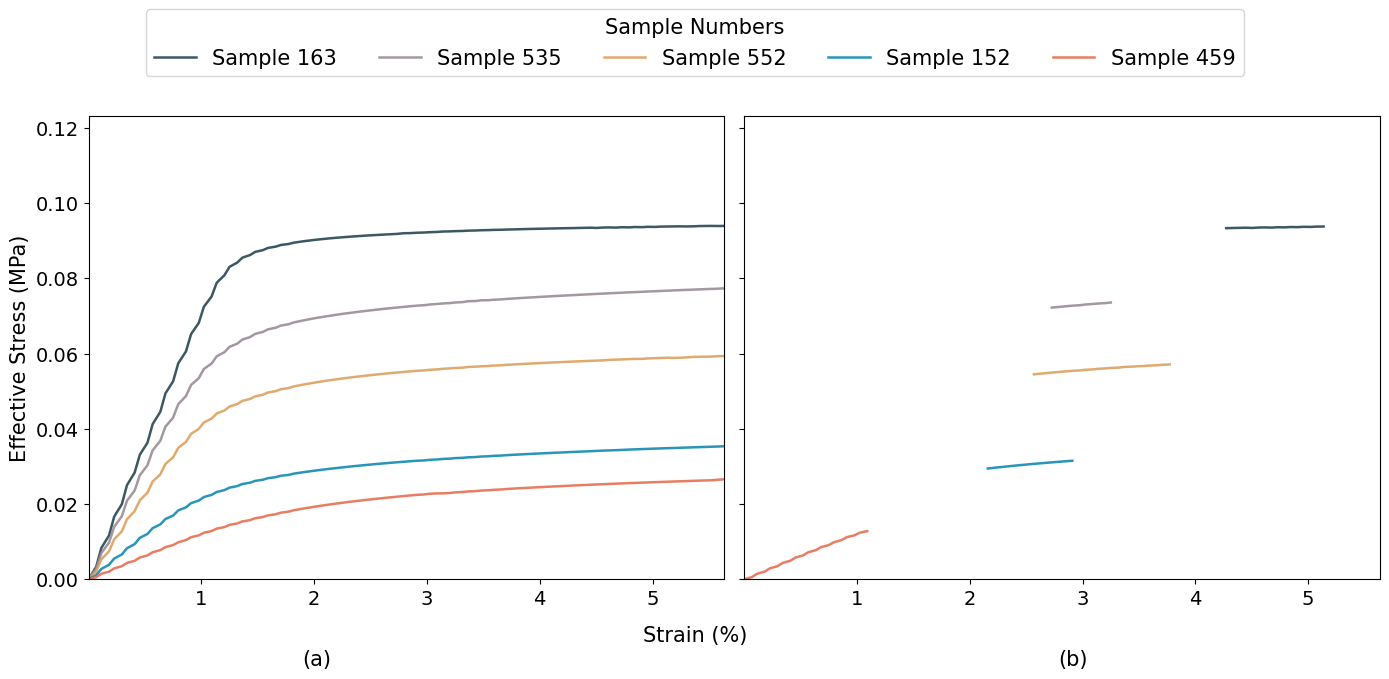}
\caption{Five randomly selected examples of stress-strain curve results from the 654 different microstructures \cite{peloquin2023neural,Lindqwister2023}: (a) original data and (b) masked partial data.} 
\end{figure}

Importantly, Guevel et al. \cite{GUEVEL2022111454} identified four microstructural features - \textit{porosity, surface area, mean curvature, and Euler characteristic} - were sufficient to capture the relationship between material microstructure and mechanical behavior under uniaxial compression. Guided by these findings, these four features were used as the target outputs for the prediction. To visualize the four features in this study, \textbf{Figure 2} presents one example microstructure of a Beech wood and the properties derived from the dataset published by Peloquin et al. \cite{peloquin2023neural}, which served as the basis for the present analysis. Porosity represents the proportion of voids relative solid based on the total volume, indicating the openness of the structure. Surface area corresponds to the boundary length between voids and solids, reflecting edge complexity. Mean curvature captures the smoothness or roughness of the edges by describing the curvature of the boundaries. Lastly, the Euler characteristic, a topological metric, quantifies the connectivity of the structure by accounting for the number of voids compared to discreet (unconnected) enclosed voids. By focusing on these four key features, the inversion process became more efficient and accurate, as the models were trained to predict the microstructural aspects most relevant to understanding material behavior. Before applying these four features, a pre-evaluation step was performed to ensure data quality. Outliers in the original dataset were identified using the Interquartile Range (IQR) method \cite{10.1007/978-981-10-7563-6_53}, a widely accepted statistical approach used to detect values significantly deviating from the majority of the data. This method computes the first quartile ($Q_1$) and third quartile ($Q_3$) for each feature, then calculates the IQR as $Q_3 - Q_1$. Any data point falling below $Q_1 - 1.5 \times \mathrm{IQR}$ or above $Q_3 + 1.5 \times \mathrm{IQR}$ is considered an outlier. As this approach does not assume a specific data distribution, it is well-suited for detecting anomalies in structural features such as porosity, surface area, mean curvature, and Euler characteristic. These outliers are visually highlighted in the plots with red markers for clarity (see \textbf{Figure 3}). To ensure a clean and consistent dataset, an additional filtering strategy was adopted: if a sample exhibited an outlier in any one of the four features, the entire sample was removed. As a result, 65 samples were discarded, and the remaining clean dataset (now 589 samples) was used for model training.

\begin{figure}[!htb]
\centering
\includegraphics[width=1\textwidth]{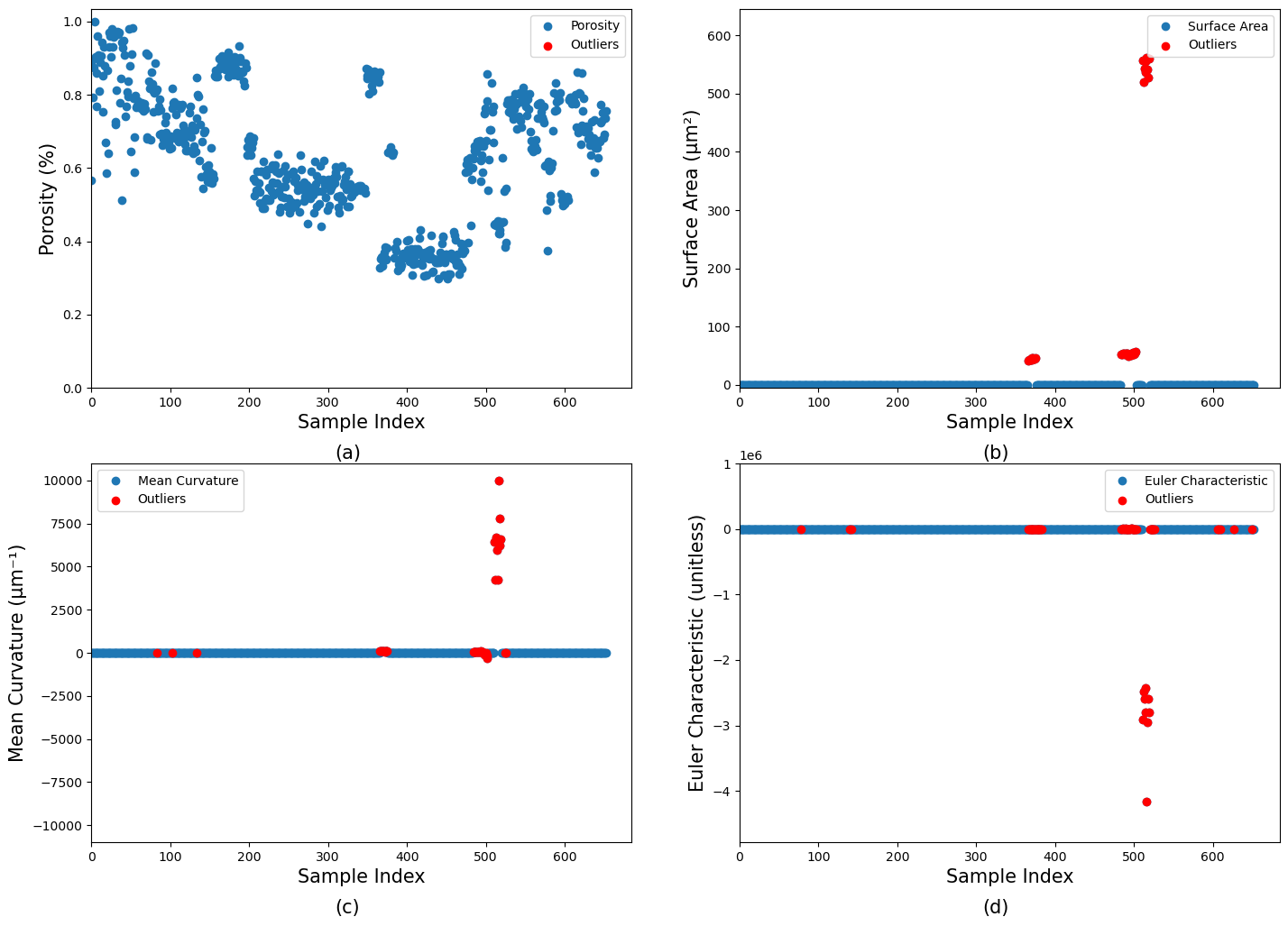}
\caption{Original data before outliers were removed: (a) porosity; (b) surface area; (c) mean curvature; (d) Euler characteristic.}
\end{figure}

 \begin{figure}[htbp]
\centering
\includegraphics[width=1\textwidth]{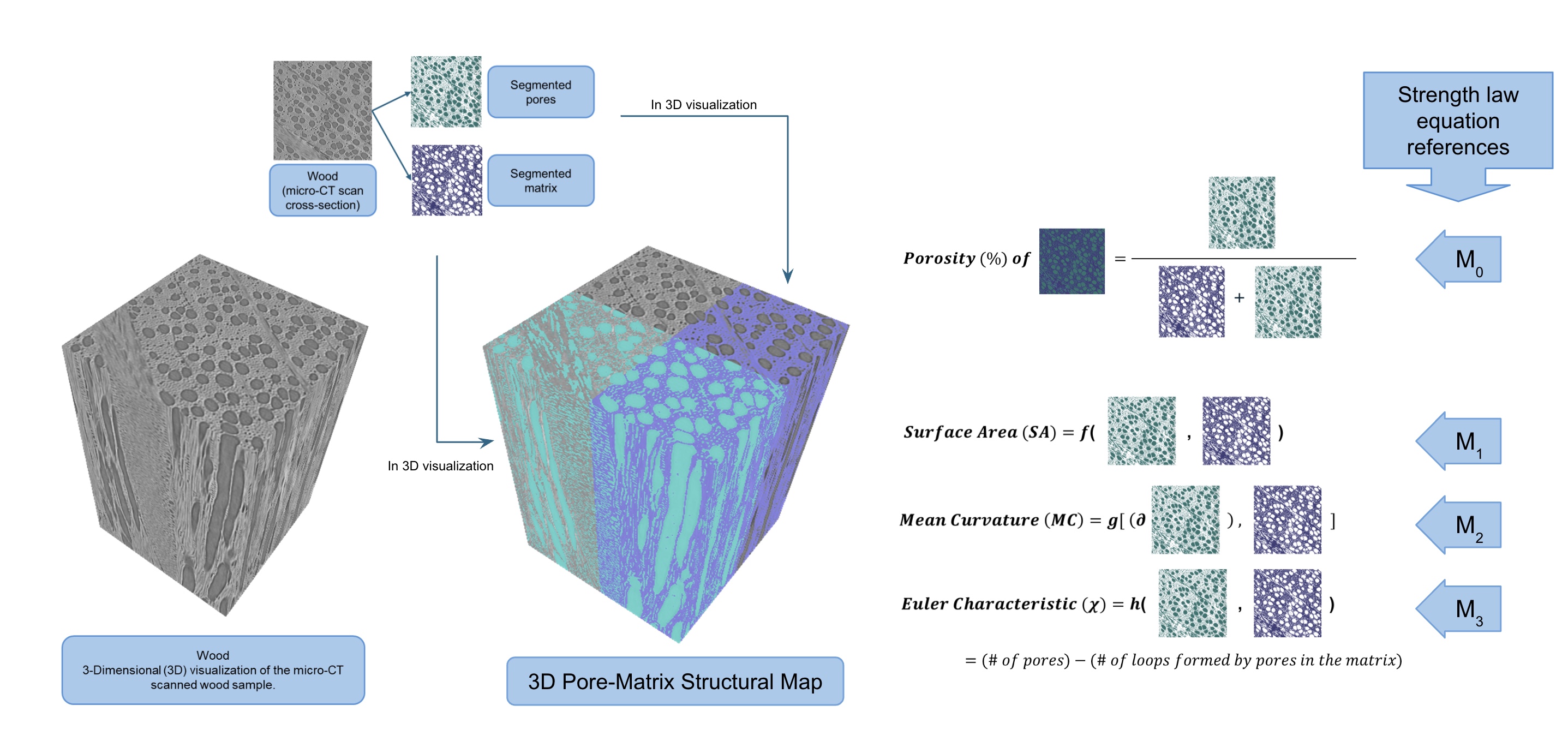}
\caption{Example microstructure \cite{peloquin2023neural} to generate stress-strain curves in previous work \cite{Lindqwister2023} and images showing each of the four main functionals used in this work.}
\end{figure}

\subsection{Models}

The LLH workflow consisted of two steps: (1) stress-strain curve reconstruction and (2) microstructure prediction (illustrated in \textbf{Figure 4}). The objective of this workflow was to first reconstruct the complete stress-strain response from partial masked data and subsequently utilize the reconstructed data to evaluate the feasibility of DL/ML models for microstructure prediction. The masked data contained only partial stress-strain information from the full curve data\cite{Lindqwister2023} and required an interpolation or prediction process to reconstruct the complete curve. This reconstruction was performed using data-driven approaches to ensure the inferred stress-strain behavior aligned with the underlying material response. Once the complete stress-strain curves were obtained, the microstructure prediction step followed. In this stage, various DL/ML methods were applied to assess the effectiveness of different predictive models in establishing a relationship between the mechanical properties and the underlying microstructure. By integrating these two steps, the LLH workflow aimed to enhance both the accuracy of stress-strain curve reconstruction and the predictive capability of DL/ML models in microstructure characterization. The following subsections provide a detailed discussion of each step in the workflow. In this study, all codes were run in Python using an NVIDIA RTX 4090 GPU.

\begin{figure}[htbp]
\centering
\includegraphics[width=1\textwidth]{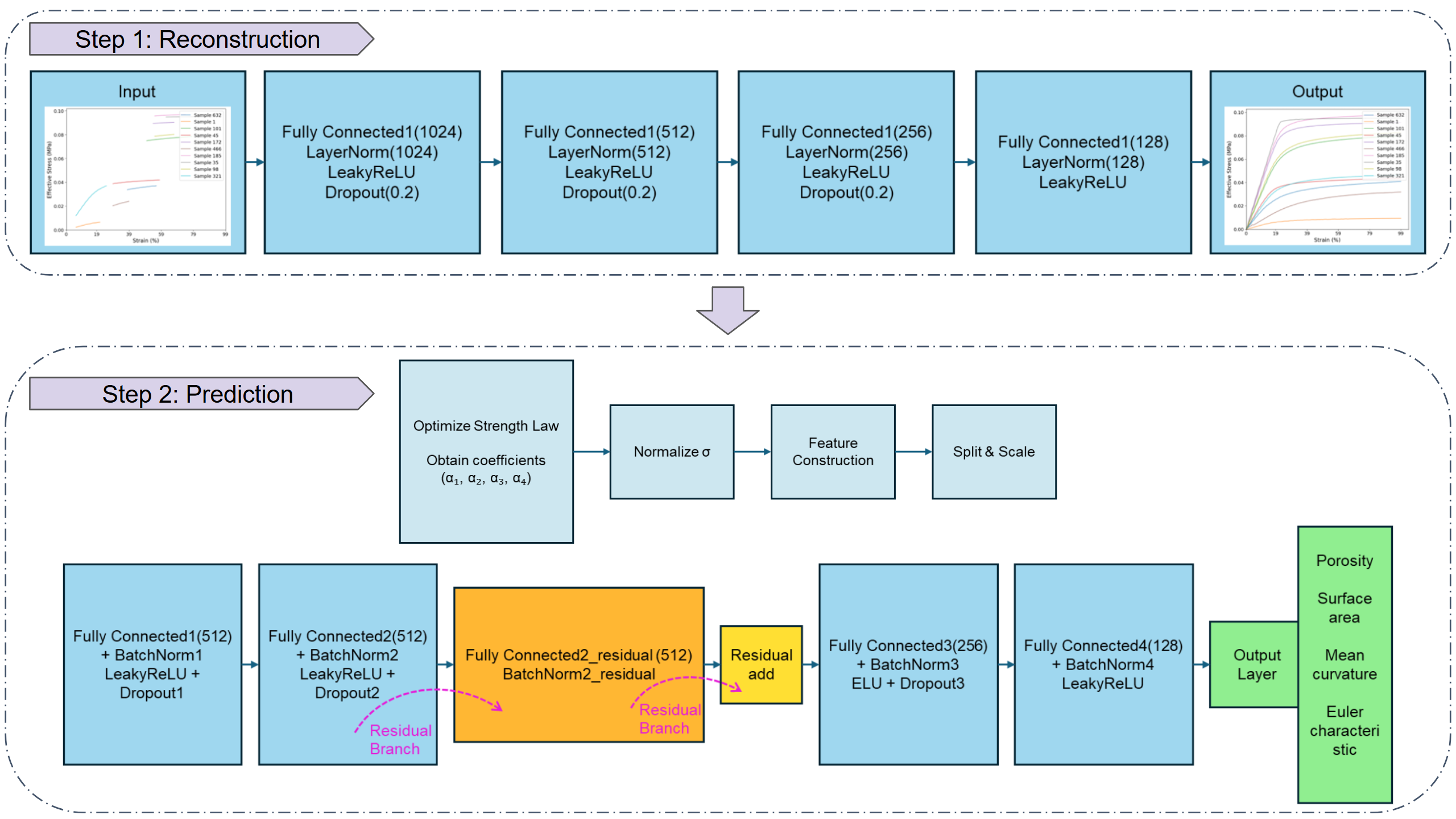}
\caption{The architecture of the LLH workflow model.} 
\end{figure}

\subsubsection{Stress-Strain Curve Reconstruction}

In the LLH framework, the first step focused on reconstructing the full stress-strain curve from partial data, to improve the accuracy of characterizing mechanical behavior before microstructural predictions. This problem was addressed using an optimized fully connected DNN, which learned the underlying nonlinear relationships between incomplete and full stress-strain responses. The reconstruction process was used to ensure the missing mechanical information was inferred effectively to enable downstream tasks to operate on a complete dataset. The input dataset consisted of masked stress-strain curves, where a portion of the stress-strain response was missing, and the corresponding complete stress-strain curves, which served as the ground truth. To ensure robust model training, several preprocessing steps were applied. First, missing values in the input data were filled using mean imputation \cite{JAMSHIDIAN200721} to fill in incomplete sections while preserving the overall distribution. Second, the data was scaled using MinMax normalization \cite{patro2015norm} to verify all features fell within a fixed range, and prevent any feature from dominating the learning process and improving numerical stability. Finally, the dataset was split into training (80 \%), validation (10 \%), and test (10 \%) sets to facilitate model training, hyperparameter tuning, and final evaluation. The validation set was used to monitor model performance and adjust hyperparameters, while the test set assessed the generalization capability of the trained model on unseen data.

The model was structured as a fully connected neural network with multiple layers designed to extract and refine features from the input data. The input layer took in the preprocessed masked stress-strain data, which was passed through a series of fully connected hidden layers with neuron sizes 1024, 512, 256, and 128, progressively reducing dimensionality while retaining critical stress-strain features. Each hidden layer was followed by layer normalization, which stabilized training by reducing internal covariate shifts, ensuring more consistent and efficient learning. The model employs LeakyReLU activation \cite{Ansel_PyTorch_2_Faster_2024} with a negative slope coefficient of 0.01 in all layers, allowing for better gradient flow and preventing vanishing gradient issues. To enhance generalization and prevent overfitting, dropout regularization with a probability of 20 \% was applied after each hidden layer, randomly deactivating neurons to force the network to learn more robust features. Finally, the output layer reconstructed the full stress-strain curve, effectively inferring the missing sections from the partial input. The model was trained using the mean squared error (MSE) loss function, which measures the discrepancy between the predicted and actual full stress-strain curves. Optimization was performed using the Adam optimizer \cite{kingma2017adammethodstochasticoptimization}, with a learning rate of 0.0003 and weight decay of $1\times10^{-4}$, ensuring stable convergence while preventing overfitting. Additionally, a step learning rate scheduler was implemented to reduce the learning rate by a factor of 0.5 at every 50 epochs, allowing finer adjustments as training progressed. The model was trained for 500 epochs using a batch size of 32, balancing training efficiency and performance.  

Model evaluation was conducted using the coefficient of determination ($R^2$ score), providing a quantitative assessment of reconstruction accuracy. Once trained, the model was applied to reconstruct full stress-strain curves from new, unseen masked inputs, effectively recovering missing mechanical information. These reconstructed curves then served as inputs for the second stage of the LLH framework, where four key microstructural features were predicted based on the recovered mechanical response.


\subsubsection{Microstructure Prediction}

To perform the inversion of predicting microstructural properties from stress-strain data, five DL/ML models were selected to test the unique strengths and suitability of the input data in the second step of the LLH model. The input dataset was structured such that each row was a set of data points from a stress-strain curve while the corresponding output reflected specific microstructural properties. This organization enabled the models to learn complex relations between stress-strain behavior and microstructural features directly. The dataset was divided into training, validation, and test subsets according to a split of 90 \% for training, 5 \% for validation, and 5 \% for testing, ensuring sufficient data for model training while providing appropriate amounts for validation and performance evaluation. Unlike the first stage, where stress-strain curves were reconstructed using a fully connected DNN, the second stage focuses on predicting microstructural features from reconstructed stress-strain data. The models employed for this task include CNN, KNN, LSTM networks, RF, and XGBoost. These DL/ML models were chosen to handle different data structures while leveraging different abilities of each method, ranging from sequence-based deep learning (LSTM) to tree-based ensemble learning (RF and XGBoost) to allow for a comprehensive evaluation of different machine learning paradigms in microstructural inference.  
A brief summary of each of the five DL/ML models are included below:

\begin{itemize}
    
\item \textbf{CNN} \cite{li2021survey, wu2017introduction} was chosen to detect spatial patterns and extract features from structured data. For this study, CNNs were adapted to utilize stress-strain data to identify local dependencies or patterns which may be predictive of microstructural properties.

\item \textbf{LSTM}, a type of recurrent neural network \cite{graves2012long, sherstinsky2020fundamentals}, was selected to test its ability to work effectively with sequential data. Given stress-strain curves were inherently sequential, LSTM was particularly suited to capture temporal dependencies in the data and provide a nuanced understanding of how input variations affect microstructural outcomes. 

\item \textbf{KNN} \cite{10.5120/3836-5332, zhang2016introduction} was included as a straightforward, instance-based learning method to predict outcomes based on proximity in the feature space. While simple, KNN serves as a benchmark for understanding how well the microstructure can be predicted based on feature similarity alone, without relying on complex learned patterns.

\item \textbf{RF} \cite{breiman2001random} was chosen to test the robust tree-based learning method and its ability to handle complex and nonlinear relationships between input features and outputs. Its ability to reduce overfitting by averaging over multiple decision trees ensures reliability, making it a solid baseline for comparison.

\item \textbf{XGBoost} \cite{chen2015xgboost, 10.1145/2939672.2939785} was selected for its ability to iteratively combine weak learners and focus on reducing prediction errors. It excels at capturing subtle relationships between input features, making it highly effective in tasks where nuanced patterns significantly impact outcomes. Its regularization techniques also help prevent overfitting to ensure the model generalizes well to unseen data. 
\end{itemize}

Each model was trained on the reconstructed full stress-strain curves as input, with the corresponding microstructural properties as output labels. The training objective was to minimize prediction error, allowing the models to extract latent patterns in stress-strain behavior which correlate with microstructural attributes. Performance was assessed using standard evaluation metrics, including $R^2$ value, to ensure accurate and robust predictions. The impact of incorporating domain knowledge, particularly through the reconstructed stress-strain curves from the first stage, was also examined by comparing model performance with and without the reconstructed data. By integrating these two stages, stress-strain curve reconstruction and microstructural prediction, the LLH framework effectively bridges mechanical response and microstructural inference, offering a data-driven approach to material characterization. This two-step methodology enhances predictive accuracy by leveraging domain knowledge in mechanical behavior while providing a robust framework for material property prediction in data-limited scenarios. 

Following the work of Guevel et al. \cite{GUEVEL2022111454}, which highlights the relationship between material strength and essential microstructural features using Minkowski functionals, the functional relationship in the present inverse problem was based on Hadwiger’s theorem, where the strength of porous materials was linked to microstructural attributes through the strength law shown in \textbf{Equation 1}:

\begin{equation}
\sigma = \sigma^* e^{\alpha_1 M_0 + \alpha_2 M_1 + \alpha_3 M_2 + \alpha_4 M_3}
\label{eq:function1}
\end{equation}

where \( \sigma \) represents the stress corresponding to each set of input features, with \( \sigma^* \) serving as a reference stress value set to 1. The parameters \( M_0, M_1, M_2, \) and \( M_3 \) correspond to porosity, surface area, mean curvature, and Euler characteristic, respectively, while the coefficients \( \alpha_1, \alpha_2, \alpha_3, \) and \( \alpha_4 \) were derived empirically from the dataset. Specifically, these coefficients were optimized in an initial step using the dataset from Lindqwister et al. \cite{Lindqwister2023}, which contains the four extracted features obtained from X-ray micro-CT scanned data. The micro-CT data consist of material samples scanned using a micro-CT machine, where the scan slices were analyzed and processed to extract the features. The strength law equation guided the optimization process, where \( M_0 \) to \( M_3 \) served as the input features. The optimized values of \( \alpha_1 \) to \( \alpha_4 \) were then substituted into the equation, yielding an updated strength law formulation. During the pretraining step, the refined equation was used to adjust the feature weights, allowing the model to learn which features should be given higher importance. This step was introduced to ensure the model effectively captured the relationship between the microstructural features and the strength of the material. These features were integrated into the inversion model to enhance the understanding of the relationship between input variables and the predictive output.

Building upon the insights gained from the comparative analyses, a specialized DNN leveraging the optimized strength law was developed. Specifically, coefficients within the strength law were optimized by minimizing the squared difference between predicted and actual log strength values obtained from micro-CT scanned samples. The resulting strength predictions were then normalized to the [0,1] range for numerical stability. These normalized predictions, combined with base features from previous neural network predictions and 35 additional numerical features, formed the dataset used for training the model. The dataset was partitioned into 80 \% training and 20 \% testing subsets, standardized using StandardScaler \cite{Ansel_PyTorch_2_Faster_2024}, and converted into PyTorch tensors, with DataLoader \cite{Ansel_PyTorch_2_Faster_2024} facilitating efficient mini-batch training. The designed DNN architecture consisted of four primary blocks. The first block comprised a fully connected layer of 512 neurons, followed by batch normalization, dropout regularization (20 \%), and a LeakyReLU activation \cite{Ansel_PyTorch_2_Faster_2024} function. The second block was similarly structured with another 512-neuron layer, batch normalization, 20 \% dropout, and LeakyReLU activation with an additional residual connection to improve gradient propagation and to mitigate vanishing gradients. This residual branch was composed of an auxiliary fully connected 512-neuron layer with batch normalization. The third block reduced dimensionality to 256 neurons and applied batch normalization, a 25 \% dropout, and an ELU activation \cite{Ansel_PyTorch_2_Faster_2024} function to enhance the learning of nonlinear feature interactions. The fourth block further reduced dimensionality to 128 neurons, incorporating batch normalization and another LeakyReLU activation. Finally, the network concluded with an output layer predicting the four Minkowski functionals  \( M_0, M_1, M_2, M_3\). The training utilized the AdamW optimizer \cite{Ansel_PyTorch_2_Faster_2024} with an initial learning rate of \(10^{-3}\) and weight decay of \(10^{-5}\), alongside a dynamic learning rate scheduler (ReduceLROnPlateau) and an early stopping criterion based on validation performance to prevent overfitting.

\subsection{Evaluation Metric}
To assess the performance of the models to predict microstructural properties, the coefficient of determination ($R^2$) was used as the primary evaluation metric. The $R^2$ score measures the proportion of variance in the target variable which was explained by the input features, providing a clear indicator of the predictive accuracy of the models. The formula for calculating $R^2$ is expressed as shown in \textbf{Equation 2}:

\begin{equation}
R^2 = 1 - \frac{\sum_{i=1}^n (y_i - \hat{y}_i)^2}{\sum_{i=1}^n (y_i - \bar{y})^2}
\label{eq:function2}
\end{equation}

where $y_i$ represents the actual values of the target variable, $\hat{y}_i$ represents the predicted values, $\bar{y}$ is the mean of the actual values, and $n$ is the number of observations. In this formula, the numerator, $\sum_{i=1}^n (y_i - \hat{y}_i)^2$, represents the sum of the squared residuals (errors), and the denominator, $\sum_{i=1}^n (y_i - \bar{y})^2$, represents the total variance in the actual data. An $R^2$ value of 1 indicates a perfect prediction, meaning the model explains all the variance in the data. Conversely, an $R^2$ value of 0 suggests the predictions of each model were equivalent to simply using the mean of the actual values. Using $R^2$ as the evaluation metric ensured a standardized and interpretable measure of the performance of the models to enable a direct comparison of the ability of each model to predict microstructural properties.
\end{flushleft}

\section{Results}
\begin{flushleft}
In this study, the aim was to assess the impact of incorporating the strength law equation (\textbf{Equation 1}) into an inverse modeling framework to assist in predicting the four critical material features: porosity (M$_0$), surface area (M$_1$), mean curvature (M$_2$), and Euler characteristic (M$_3$). To evaluate the influence of the strength law equation, five different predictive models were tested both with and without \textbf{Equation 1} as input information. Before evaluating the inverse prediction of microstructural features, the first stage of the LLH framework was analyzed to assess the accuracy of the stress-strain curve reconstruction. The performance of DNN was examined by comparing the reconstructed stress-strain curves with the expected ground truth data. \textbf{Figure 5} presents a direct comparison between the DNN-reconstructed and actual stress-strain curves for the same five samples shown in \textbf{Figure 1} , illustrating the ability of the model to capture key mechanical features such as yielding behavior and strain hardening effects. The results clearly demonstrate a strong correspondence between the reconstructed curves and the actual data points across multiple randomly selected samples. The ground truth curves, shown in solid, coral lines, and the predicted curves, depicted as dashed, teal lines, closely align with one another, highlighting the precision of the model in replicating the original mechanical response. In addition, the masked input data, illustrated with bold, purple lines, emphasizes the regions used for predictions. Quantitatively, the accuracy of the reconstructed stress-strain data was evaluated using the coefficient of determination, which achieved a test $R^2$ score of 0.9851 further emphasizing the robustness of the DNN model to accurately predict stress-strain behaviors.

\begin{figure}[!htb]
    \centering
    \includegraphics[width=0.8\textwidth] {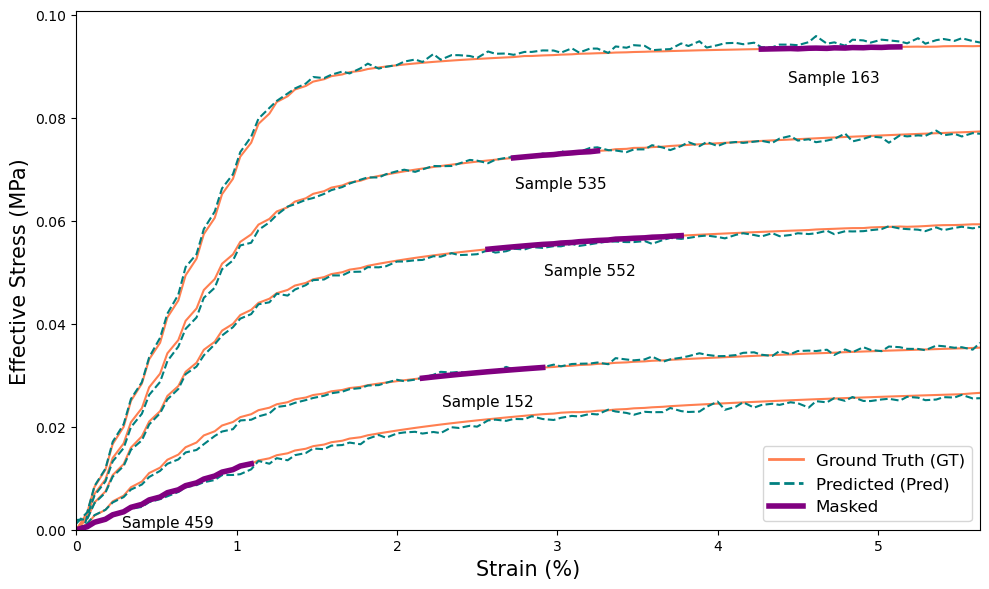}
    \caption{Representative test data and results comparing the ground truth, masked training data, and predicted (DNN-reconstructed) stress-strain curves.}
    \label{fig:reconstruction_comparison}
\end{figure}

With the reconstructed stress-strain curves established, the impact of the results on microstructural property predictions was evaluated. The second stage of the LLH framework utilized these reconstructed stress-strain curves as input to predict microstructural properties to assess how well the inferred mechanical response enhanced predictive accuracy. Model performance was assessed using the $R^2$ score as a quantitative metric to determine how closely the predicted microstructural features aligned with the ground truth values. This analysis highlights the extent to which incorporating the strength law equation and reconstructed mechanical response improved predictive accuracy within the inverse modeling framework.

The results are summarized in \textbf{Figure 6}, which illustrates the predictions of CNN, KNN, LSTM, RF, and XGBoost for \textbf{porosity}. In this figure, the red dotted line indicates the best match, the coral-colored (light) points represent the predictions without the domain function, while the teal-colored (dark) points correspond to the predictions with the domain function. \textbf{Figure 6} presents a comparison of the porosity predictions with and without the incorporation of the stress function during model training. The surface area, mean curvature, and Euler characteristic results are included in \textit{Supplementary Material}. 

\begin{figure}[!htb]
\centering
\includegraphics[width=1\textwidth]{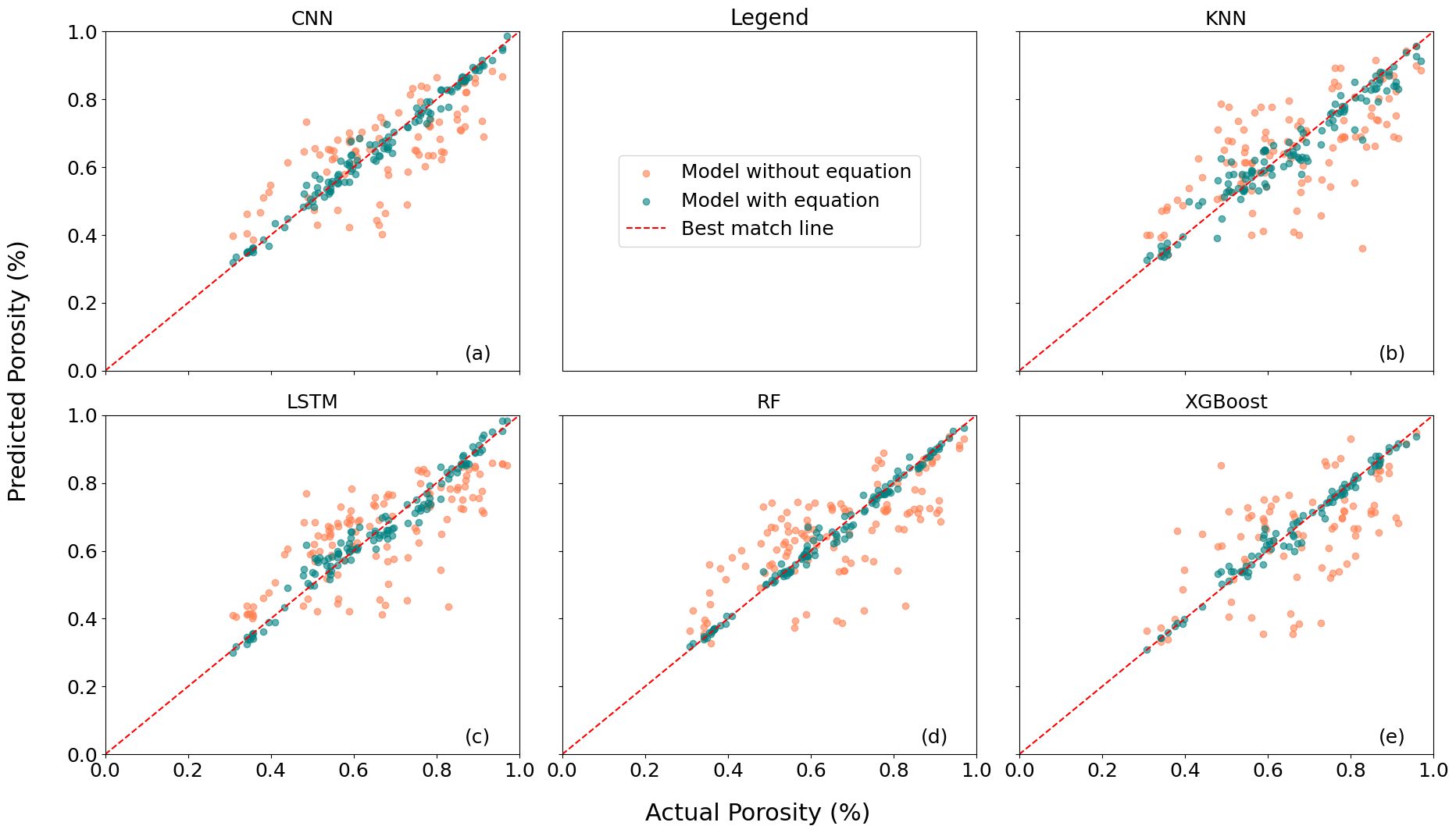}
\caption{Comparison of the predicted porosity values from the five models with and without the inclusion of the strength law equation: (a) CNN, (b) KNN, (c) LSTM, (d) RF, and (e) XGBoost.}
\end{figure}

A more straightforward summary is in \textbf{Table 1} including the runtime. The coefficient of determination ($R^2$) is calculated as an average across all microstructural predictions, providing an overall assessment of model performance. The $R^2$ values indicate the potential for improved predictive accuracy when the stress function is included in the model.

\begin{table}[!htb]
    \centering
    \caption{Comparison of R\textsuperscript{2} and runtime for different models with and without domain knowledge.}
    \renewcommand{\arraystretch}{1.2}
    \setlength{\arrayrulewidth}{0.5mm}
    \setlength{\tabcolsep}{10pt}
    \begin{tabular}{lcc|cc}
        \toprule
        \textbf{Model} &
        \multicolumn{2}{c|}{\textbf{Without Function}} &
        \multicolumn{2}{c}{\textbf{With Function}} \\
        \cmidrule(r){2-3} \cmidrule(r){4-5}
        & R\textsuperscript{2} & Runtime (seconds) 
        & R\textsuperscript{2} & Runtime (seconds) \\
        \midrule
        CNN      & 0.2500 & 0.8   & 0.5617 & 7.5 \\
        KNN      & 0.1456 & 1.1   & 0.4129 & 1.2  \\
        LSTM     & 0.2879 & 1.7   & 0.4867 & 2.2  \\
        RF        & 0.1866 & 8.3   & 0.6240 & 2.6 \\
        XGBoost  & -0.0894 & 1.7   & 0.4619 & 3.9 \\
        \bottomrule
    \end{tabular}
    \label{tab:r2_runtime_comparisons_clear}
\end{table}


\section{Discussion and Conclusions}
The objective of this study was to utilize outputs from an existing forward problem combined with prior knowledge as inputs into DL/ML models to predict microstructural features. Specifically, the model developed in the present study termed Learning Latent Hardening (LLH) was designed to first reconstruct complete stress-strain curve responses (i.e., mechanical response under uniaxial compressive loading) from partial segments of the actual (ground truth) stress-strain data and then utilize the reconstructed curves to predict four key microstructural features of porous material microstructures \cite{GUEVEL2022111454, Lindqwister2023}: porosity (\(M_0 \)), surface area (\( M_1 \)), mean curvature (\( M_2 \)), and Euler characteristic (\( M_3 \)). 

Driven by the encouraging results presented in the previous sections, DNN was explored to improve microstructure prediction. \textbf{Figure 7} includes a comparison of the prediction results obtained using the DNN and the RF models for the four key structural parameters (results without domain knowledge are included in \textit{Supplementary Material}). Across all examined properties, the DNN outperformed the RF model. Specifically, the predictions from the DNN closely follow the ground truth reference, suggesting a robust ability to accurately capture complex relationships within the data. For example, the surface area and mean curvature predictions from the DNN exhibit less deviation from the ground truth, aligning closely with the central reference line, whereas the RF predictions display greater scatter and reduced accuracy. These results highlight the robustness and effectiveness of the DNN model in capturing intricate patterns and delivering reliable predictions across all the evaluated structural parameters.
\end{flushleft}

\begin{figure}[!htb]
\centering
\includegraphics[width=\textwidth]{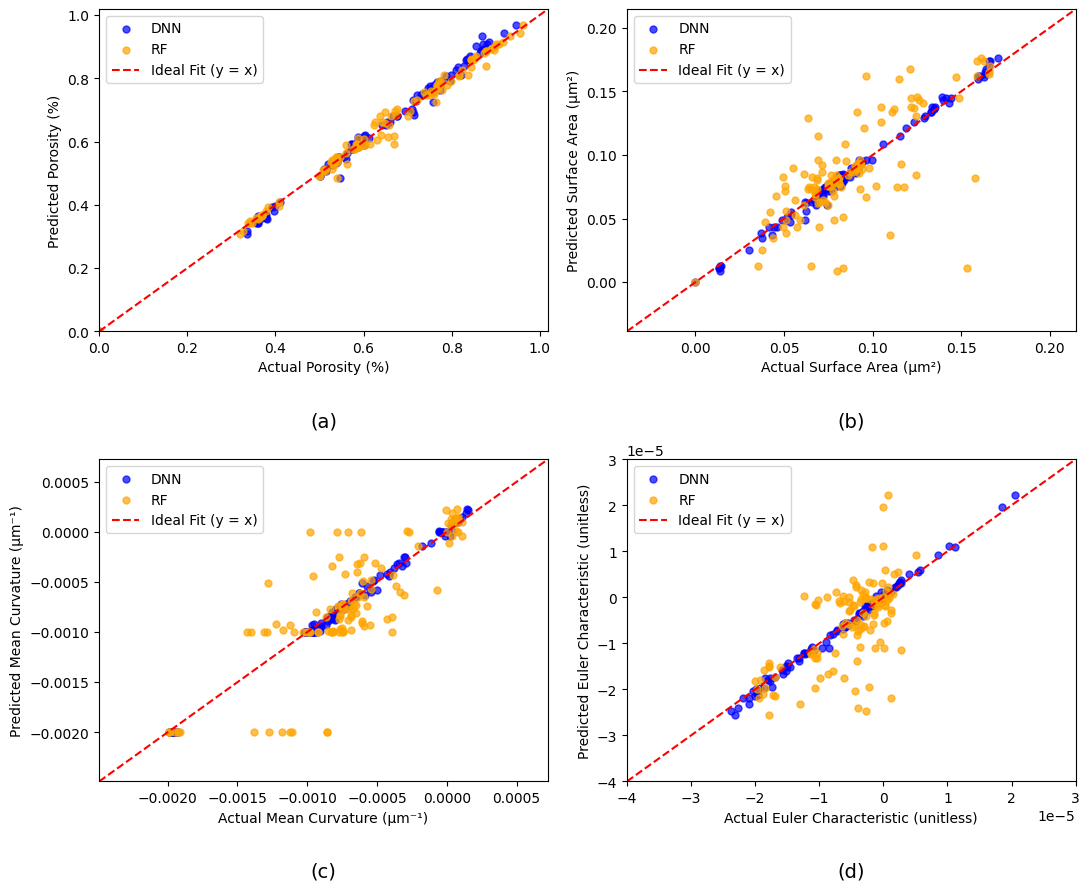}
\caption{Comparison of prediction results between the DNN and RF models for different parameters: (a) porosity, (b) surface area, (c) mean curvature, and (d) Euler characteristic.} 
\end{figure}
\vspace{-5mm}
\begin{flushleft} 
The improved performance of the DNN compared to the RF model arises primarily from the inherent ability of DNN to handle highly nonlinear relationships within stress-strain data. Unlike RF, which relies on discrete splits and hierarchical decision boundaries, the DNN leverages multilayered architectures which can more effectively capture subtle patterns and continuous feature interactions present in data such as the complex stress-strain curves used in the present study. Consequently, the DNN approach provides superior generalization and accuracy (R$^2$ value of 0.9936 as shown in \textbf{Table 2}), particularly when predicting intricate microstructural properties which depend on subtle variations in mechanical response data. This inherent capability makes the DNN beneficial for applications requiring detailed characterization and precise predictions of material behavior from stress-strain relationships, further supporting its application in advanced material science, design, and related fields.
\end{flushleft}

\begin{table}[!htb]
    \centering
    \caption{R\textsuperscript{2} and runtime for DNN model with and without domain knowledge.}
    \renewcommand{\arraystretch}{1.2}
    \setlength{\arrayrulewidth}{0.5mm}
    \setlength{\tabcolsep}{10pt}
    \begin{tabular}{lcc|cc}
        \toprule
        \textbf{Model} &
        \multicolumn{2}{c|}{\textbf{Without Function}} &
        \multicolumn{2}{c}{\textbf{With Function}} \\
        \cmidrule(r){2-3} \cmidrule(r){4-5}
        & R\textsuperscript{2} & Runtime (seconds) 
        & R\textsuperscript{2} & Runtime (seconds) \\
        \midrule
        DNN      & 0.1005 & 2.0   & 0.9936 & 7.7 \\

        \bottomrule
    \end{tabular}
    \label{tab:r2_runtime_comparison_clear}
\end{table}

\vspace{-5mm}
\begin{flushleft}
In conclusion, incorporating domain knowledge improved the prediction of microstructural properties from stress-strain data, leading to notably higher R$^2$ values compared to models lacking such knowledge. Specifically, the domain-informed models demonstrated a superior ability to accurately capture underlying material behaviors, as evidenced by better alignment between predicted and actual values in the evaluation metrics, supported by visualizations that showed reduced prediction error. The findings demonstrate when domain-specific information is intergated to guide the model the performance of DL/ML  models is enhanced, particularly in fields like geomechanics and material engineering, where the underlying physical relationships are complex and governed by the laws of physics. By bridging the gap between DL/ML and material science, this approach shows the potential to uncover refined relationships between mechanical behavior and microstructural features, ultimately improving model accuracy and interoperability. The capability of LLH to accurately reconstruct and subsequently utilize the stress-strain behavior further emphasizes the strength of this integrated approach. The integration of domain-informed features not only improved the predictive power of the models but also demonstrates how these models can be applied to practical engineering problems. The results underscore the importance of a multidisciplinary approach combining computational techniques with foundational knowledge of material behavior. 

\end{flushleft}

\begin{flushleft}\textbf{Funding}\\
The Duke University Department of Civil and Environmental Engineering is greatly acknowledged for their support. 
\end{flushleft}

\begin{flushleft}\textbf{Acknowledgments}\\
Special thanks are extended to Dr. Hossein (Amir) Salahshoor for insightful discussions and for sharing ideas that contributed to this work.
\end{flushleft}

\begin{flushleft}\textbf{Data and Code Availability}\\
The dataset is published and openly available to interested researchers: \cite{peloquin2023neural}.

The Learning Latent Hardening (LLH\_model) code is available on GitHub at: 

DOI: 10.5281/zenodo.15156211.

\end{flushleft}

\bibliographystyle{unsrtnat}
\bibliography{ref1119}

\section{Supplementary Materials}

Here attached are the comparisons of the prediction results between models with and without function.

\begin{figure}[htbp]
\centering
\includegraphics[width=\textwidth]{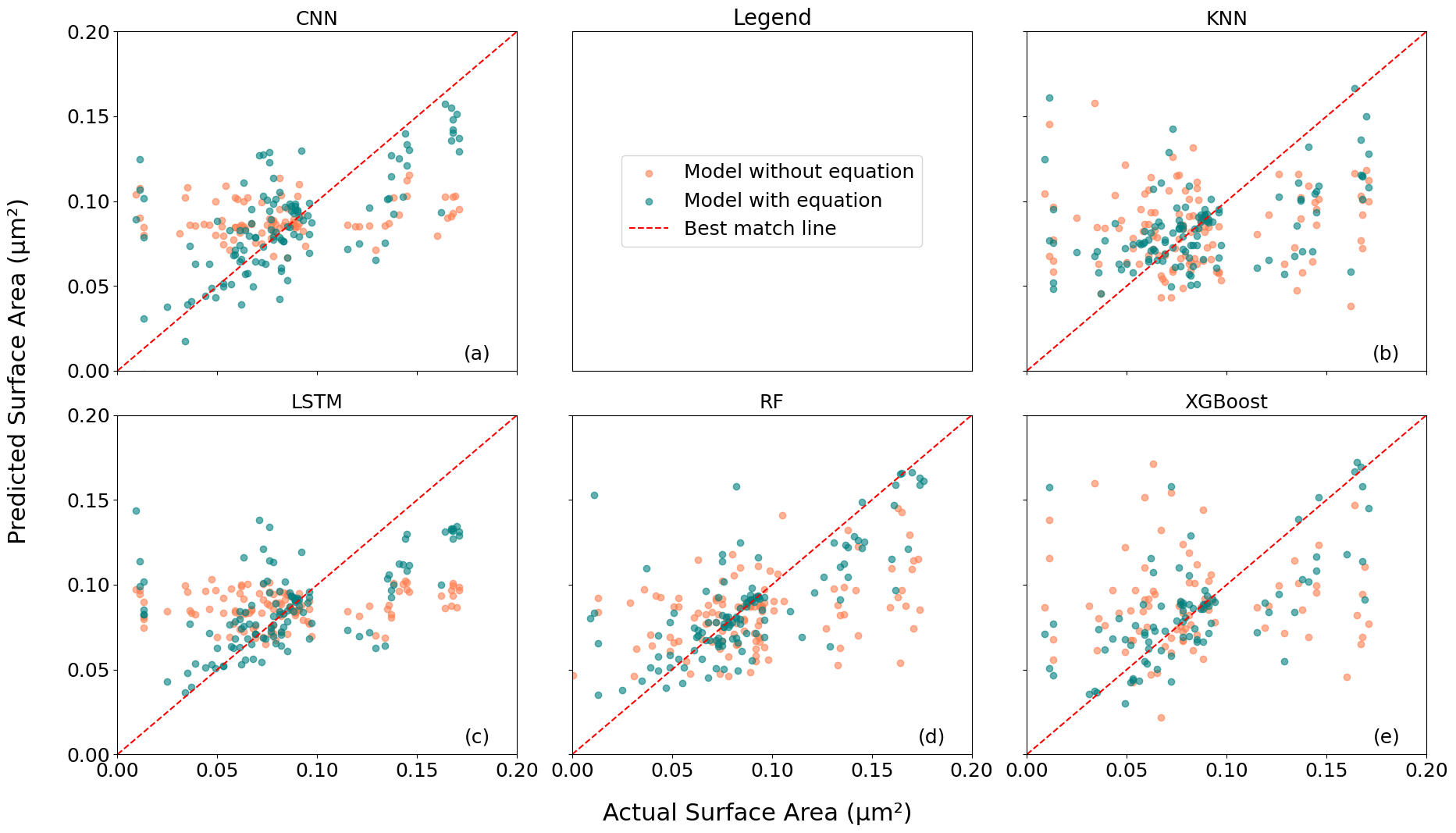}
\caption{Comparison of the predicted surface area values from the five models with and without the inclusion of the strength law equation: (a) CNN, (b) KNN, (c) LSTM, (d) RF, and (e) XGBoost. }
\end{figure}

\begin{figure}[htbp]
\centering
\includegraphics[width=\textwidth]{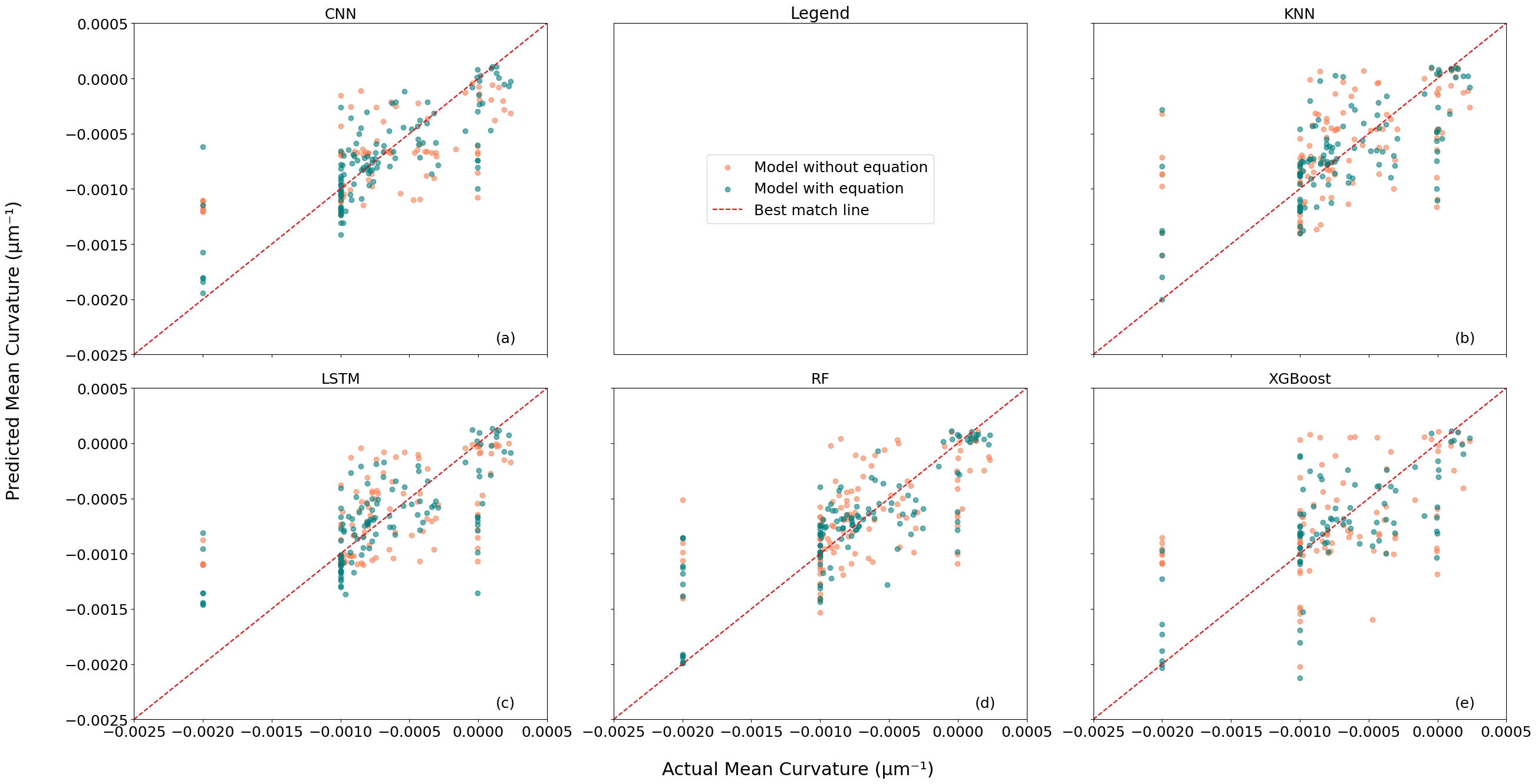}
\caption{Comparison of the predicted mean curvature values from the five models with and without the inclusion of the strength law equation: (a) CNN, (b) KNN, (c) LSTM, (d) RF, and (e) XGBoost.} 
\end{figure}

\begin{figure}[htbp]
\centering
\includegraphics[width=\textwidth]{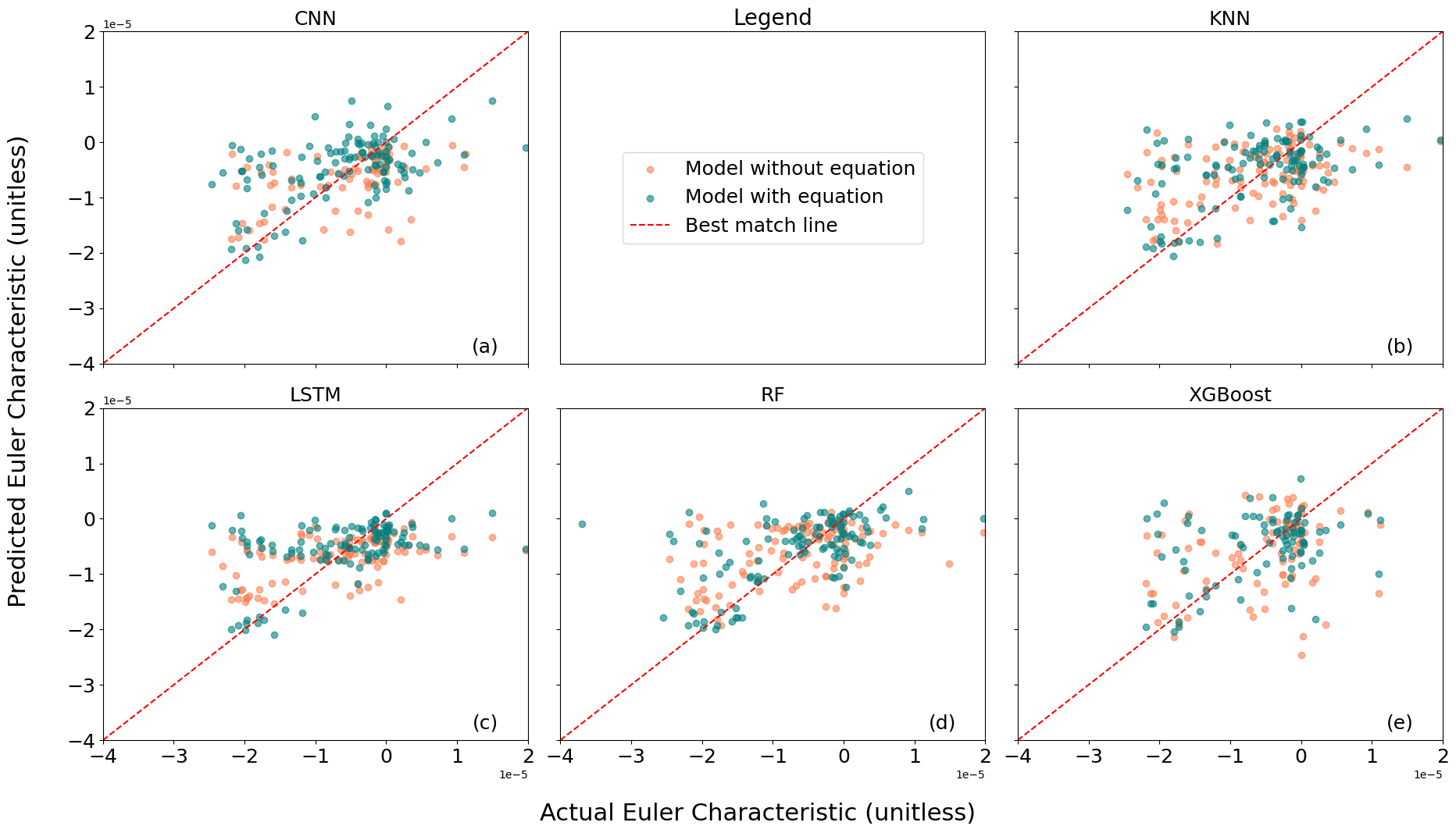}
\caption{Comparison of the predicted Euler Characteristic values from the five models with and without the inclusion of the strength law equation: (a) CNN, (b) KNN, (c) LSTM, (d) RF, and (e) XGBoost.} 
\end{figure}

\begin{figure}[htbp]
\centering
\includegraphics[width=\textwidth]{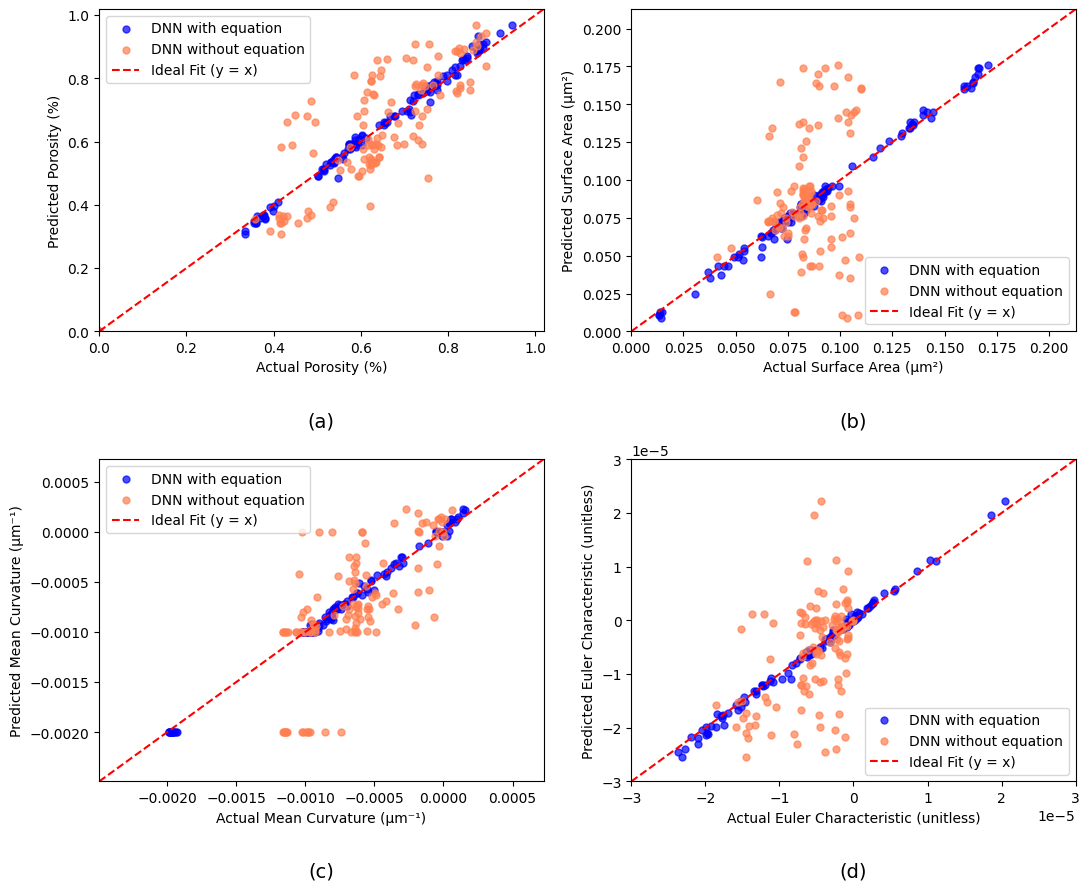}
\caption{Comparison of the performances between DNN with equation and DNN without equation.  }
\end{figure}

\end{document}